\documentclass[twoside]{deon16}

\usepackage{deon16macro}

\usepackage{graphicx}
\usepackage{amsmath}
\usepackage{amssymb}
\usepackage{float} 
\usepackage[table,dvipsnames]{xcolor}
\usepackage{todonotes}

\usepackage[most]{tcolorbox}
\usepackage{graphicx}
\usepackage{subcaption}

\usepackage{soul}
\usepackage{placeins} 

\usepackage{ltl}
\usepackage{habbas-macros}

 \newcommand{\mdp}{\mathcal{M}}

\newcommand{\mcmdp}[1]{\mdp||#1}
\def\mcmdpi{\mcmdp{\pi}}
\def\pipo{\pi^{\mathsf{PO}}}
\def\Exp{\mathbb{E}}
\def\eauformula{\psi}
\def\pctlstate{\formula}
\def\pctlpath{\varphi}

\def\lastname{Makarova and Abbas}
\usepackage{hyperref}
\begin{document}

\begin{frontmatter}
  \title{Deontically Constrained Policy Improvement in Reinforcement Learning Agents}
  \author{Alena Makarova}\footnote{makarova@oregonstate.edu. Research partially supported by NSF awards 2145291 and 2416461.}
  \author{Houssam Abbas}
  \address{School of Electrical Engineering and Computer Science  \\ Oregon State University  \\ Corvallis, OR}

  \begin{abstract}
  Markov Decision Processes (MDPs) are the most common model for decision making under uncertainty in the Machine Learning community. An MDP captures non-determinism, probabilistic uncertainty, and an explicit model of action. 
  A Reinforcement Learning (RL) agent learns to act in an MDP by maximizing a utility function.
  This paper considers the problem of learning a decision policy that maximizes utility subject to satisfying a constraint expressed in deontic logic.

  In this setup, the utility captures the agent's mission - such as going quickly from A to B. The deontic formula represents (ethical, social, situational) constraints on how the agent might achieve its mission by prohibiting classes of behaviors.
  We use the logic of Expected Act Utilitarianism, a probabilistic stit logic that can be interpreted over controlled MDPs.
  We develop a variation on policy improvement, and show that it reaches a constrained local maximum of the mission utility.
  Given that in stit logic, an agent's duty is derived from value maximization, this can be seen as a way of acting to simultaneously maximize two value functions, one of which is implicit, in a bi-level structure. 
  We illustrate these results with experiments on sample MDPs.
  \end{abstract}

  \begin{keyword}
  expected act utilitarianism, Markov decision process, reinforcement learning, policy improvement, stit logic
  \end{keyword}
\end{frontmatter}

\section{Introduction}
\label{sec:intro}
The most widely used model for designing artificial agents that act under uncertainty is the Markov Decision Process, or MDP \cite{sutton2018reinforcement}.
In an MDP, both the agent to be designed and the environment in which it operates (including potentially other agents) are modeled using non-determinism and stochasticity.
A \textit{reinforcement learning} (RL) agent learns to act in this environment by receiving rewards along the way. The rewards are such that maximizing aggregate reward yields desired behavior. 
E.g. if the objective is to reach a point $B$, the reward might be positive at $B$, -100 at obstacles, and 0 everywhere else.
In most cases, the reward structure is not fully known ahead of time, so the RL agent discovers what it is at the same time that it tries to maximize it.

Our purpose is to place deontic constraints on a utility-maximizing RL agent. The constraints represent ethical, social, or situational norms that apply in a given domain or context. 
So the agent retains its mission, encoded as an aggregate reward to be maximized, while the norms constrain how the agent goes about accomplishing its mission by allowing and disallowing various classes of behaviors.
For example a home care robot might have a general mission to assist an elderly person in daily life activities. 
This mission is captured in the reward function.
The normative constraints might revolve around the respect of private spaces and cultural notions of decency.
Thus, our problem is one of finding a decision policy for the agent that best accomplishes the mission, out of all policies that satisfy the normative constraints.
Mathematically, a policy that accomplishes the mission is one that maximizes aggregate reward, aka \textit{utility}.
So the mathematical problem to solve is a constrained maximization: out of all policies that satisfy the (normative) constraints, find one that maximizes utility.
The technical core of this paper is a new variant of \textit{policy improvement} to take into account the normative constraints.

Our starting point for formalizing deontic constraints is the logic of Expected Act Utilitarianism, or EAU \cite{SheaBlymyer_Abbas_2024}. 
EAU is a stit logic designed for probabilistic agents, such as RL agents over MDPs.
In EAU, states of affairs in the world are formalized using statements in Probabilistic Computation Tree Logic (PCTL) \cite{baier2008principles}, which is a logic for specifying behaviors over branching time structures in the presence of stochasticity. E.g. in PCTL we can formalize statements such as `With probability at least 0.5, the agent eventually reaches the goal state $g$ and simultaneously avoids the pit $p$'. 
In EAU, we can then say that at a given evaluation index, the agent \textit{ought to see to it that}, with probability at least 0.5, it eventually reaches the goal and simultaneously avoids the pit.
Such probability thresholds might seem arbitrary from a general perspective (why not 0.51? why not 0.99?), but in engineering system design they often derive from a study of failure rates of the system's components and a broader assurance case for establishing what is acceptable in a given concept of operations.

EAU relies on the strategic stit, or more precisely, on a probabilistic extension of strategic stit which we developed in \cite{SheaBlymyer_Abbas_2024}. 
We showed how EAU formulas can be interpreted over the executions of MDPs by translating the MDP into a stit model, and evaluating the formula there.
We also showed that utility-maximizing policies in an MDP, aka \textit{optimal policies}, are precisely the optimal strategies in the probabilistic stit model translation of the MDP, with optimality based on dominating \textit{expected} utility. This allows us to leverage PCTL model-checking in the analysis of an agent's deontic obligations in the stit framework.

In this paper we extend that work to tackle constrained policy synthesis: given a utility to be maximized, and given an EAU constraint formula $\eauformula$ to be satisfied, how does one synthesize (i.e., automatically generate) a $\eauformula$-satisfying policy that maximizes the utility? We develop a policy improvement algorithm that solves this to local optimality: that is, our algorithm converges to a local, but not necessarily global, optimizer of the constrained optimization problem.

\paragraph{Related works.}

In this paper, a norm is a statement of obligation, permission or prohibition, and is represented using a formula in a deontic logic.
Prior work has sought to formulate an RL agent's \textit{mission} as a logical formula in some temporal logic, typically Linear-time Temporal Logic (LTL) \cite{djeumou23irl}, or (much less frequently) in Probabilistic Computation Tree Logic (PCTL) \cite{morteza2012}.
That is, in those works, there is no reward, only a logical formula that encodes the mission, and algorithms are given for computing a policy that produces formula-satisfying behavior(s)~\cite{Alur2022transformingspecs}. 
LTL over MDPs is used in \cite{Kasenberg_Scheutz_2018} to reason about normative conflicts; we adopt PCTL as a better suited formalism for stochastic systems, and are not (yet) concerned with normative conflicts.
Reward machines \cite{topcu21rewardmachines} encode non-Markovian rewards, but automata for representing PCTL formulas are significantly more complex \cite{pautomata10} - whether they can be translated into rewards is left for potential future work.

Constrained MDPs offer another direction: maximizing one reward signal while ensuring constraints on some other rewards~\cite{eitan04cmdps}. However, we optimize a single explicit reward while the constraint is logical, not numeric. Work such as~\cite{achiam2017constrained} on constrained policy optimization proposes gradient-based methods under expectation and risk constraints, but these techniques are not designed to enforce logical constraints like PCTL.

Recent work on $\omega$-regular constraints~\cite{rlj2024omega} uses automata to encode logical properties and optimizes cumulative reward while satisfying acceptance conditions. 

While expressive, their method assumes the constraint is linear-time. In contrast, our method handles PCTL constraints directly, without constructing product automata, and integrates constraint satisfaction into each policy improvement step.

The work \cite{ropke2024divide} tackles the multi-objective scenario. We have one objective. But the underlying techniques of \cite{ropke2024divide} might be leveraged in our setting as they also have to work with a constrained policy improvement algorithm. However, our constraints are time-variable which complicates the optimization.

In the literature on stit theory, our use of expected utility is closest to \cite{broersen21utilstit}, where conditional probability is used as a primitive to quantify over histories; their goal is belief revision. Our paper works with classical probability, connects the \textit{strategic} stit to MDPs, and the goal is policy synthesis. 
The probabilistic (x)stit logic of \cite{BROERSEN13probabilisticstit} defines a notion of action success relative to the agent's own beliefs about said success, formalized with subjective probabilities. Our stit achievement and obligation operators are standard, but the dominance relation between actions is modified.
Strategic stit, and its connections to ATL \cite{broersen06stitatl} and Coalition Logic \cite{BROERSEN06stitCL}, has been a key way of thinking about strategies of groups of agents, since a strategy to achieve something must account for other agents' actions. In this work all other agents are lumped into the non-determinism and stochastic aspects of the model.

Abel et al. \cite{Abel2016ReinforcementLA} have argued for the use of standard RL to `solve' ethical dilemmas. It is now known~\cite{skalse2022rewardhypothesis,yang2022reinforcement} (including thanks to work by some of the authors of \cite{Abel2016ReinforcementLA}) that some common goals cannot be captured by a scalar reward.
Our approach is philosophically aligned with such critiques of the reward hypothesis—the assumption that any desirable behavior can be encoded through scalar reward functions. By separating the reward (representing the mission) and the constraint (representing the norm), we show how logical constraints can complement utility optimization.

\paragraph{Contributions.} The paper's technical contribution is a new policy improvement algorithm which maximizes utility over the space of policies that satisfy a PCTL constraint. We prove convergence to local optimality. We implemented our algorithm and illustrate its operation on sample MDPs. 

\paragraph{Organization.} Technical background is presented in Section \ref{sec:background}.
Our policy improvement algorithm is presented in Section \ref{sec:policy improvement}.
Examples illustrate its operation in Section \ref{sec:experiments}, and we conclude in Section \ref{sec:conclusion}.

\section{Background on PCTL and Reinforcement Learning}
\label{sec:background}

\PCTL~\cite[Ch. 10]{baier2008principles} is a widely used branching-time logic for specifying the behavior of stochastic systems. Given a set $A$, $2^A$ denotes its powerset, $A^*$ is the set of finite sequences from $A$ and $A^\omega$ is the set of infinite sequences from $A$.
We write $x \sim f$ when random variable $x$ is drawn following distribution $f$.

\textbf{Models.}
PCTL formulas are interpreted over Markov chains.
A \textit{Marov chain} (MC) is a tuple $M=(S,\iota,AP,L,P)$ with finite state space $S$, initial probability distribution $\iota:S\rightarrow [0,1]$, set of atomic propositions $AP$, labeling function $L:S\rightarrow 2^{AP}$ and transition probability function $P:S\times S\rightarrow [0,1]$ s.t. $\sum_{s'\in S}P(s,s')=1$ for all $s$.
A \textit{path} in the chain is an infinite sequence of states $s_0s_1s_2\dots$ s.t. $s_0 \sim \iota, s_{k+1} \sim P(s_k,\cdot)$.
It is possible to associate a non-trivial probability mass with sets of infinite executions of a Markov chain through the notion of cylinder sets - see \cite{baier2008principles} and this paper's Appendix.
We henceforth assume that there exists a state $s_{init}$ s.t. $\iota(s_{init})=1$, so $s_{init}$ is the unique initial state of the MC.

\textbf{Syntax.}
PCTL \textit{state formulas} are formed according to the grammar
\[\pctlstate \defeq \top~|~p~|~\pctlstate\land \pctlstate~|~\neg \pctlstate ~|~ P_J(\pctlpath)\]
where $p$ is an atomic proposition from a finite set $AP$, $J\subseteq [0,1]$ is an interval with rational endpoints, and $\pctlpath$ is a PCTL \textit{path formula} formed according to the following grammar
\[\pctlpath \defeq X \pctlstate ~|~\pctlstate U \pctlstate~|~ \pctlstate U^{\leq n}\pctlstate\]

\begin{figure}[t]
    \centering
    \begin{subfigure}{0.45\textwidth}
        \centering
        \includegraphics[width=\textwidth]{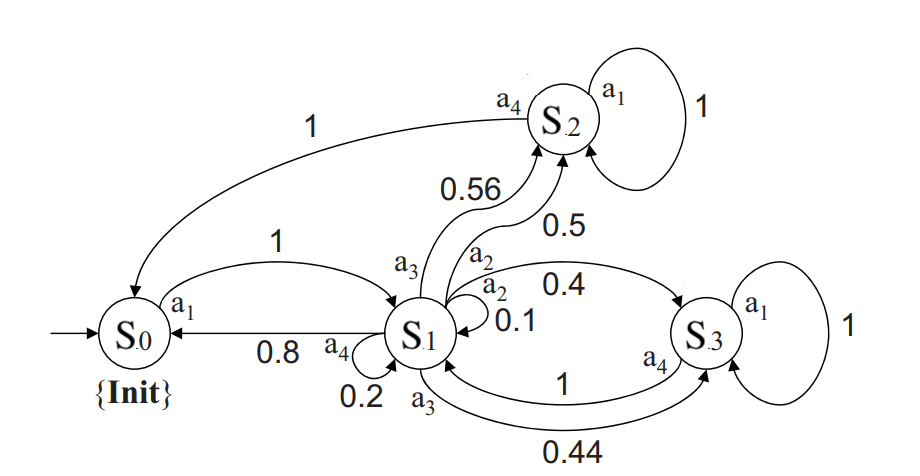}
        \caption{MDP1 from \cite{morteza2012}. 
        The initial state is $s_0$, arrows are labelled with the actions $a_n$ and transition probabilities $T(s,a,s')$. The rewards (not shown) assigned to $s_0,\dots,s_3$ are (in that order) 5, 10, 7, 3.}
    \end{subfigure}
    \hfill
    \begin{subfigure}{0.45\textwidth}
        \centering
        \includegraphics[width=\textwidth]{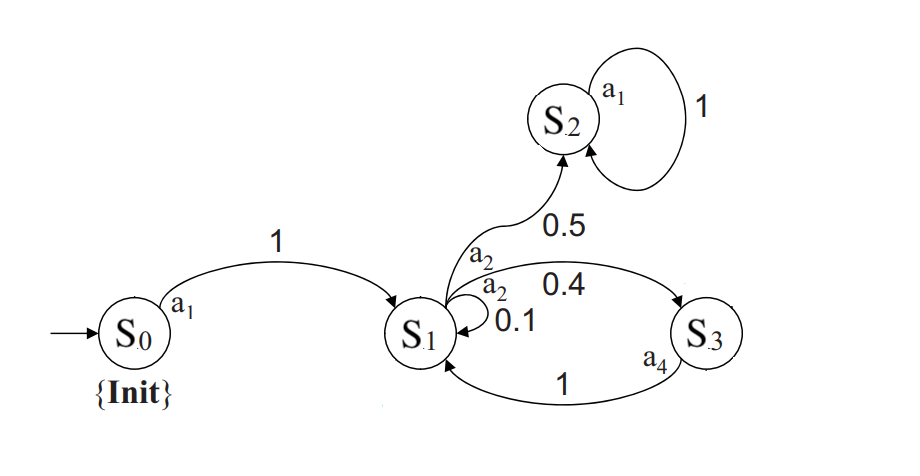}
        \caption{Markov Chain obtained by applying policy $\{s_0 \mapsto a_1, s_1 \mapsto a_2, s_2 \mapsto a_1, s_3 \mapsto a_4\}$ to MDP1.}
    \end{subfigure}
    \\
    \begin{subfigure}{0.45\textwidth}
        \centering
        \includegraphics[width=\textwidth]{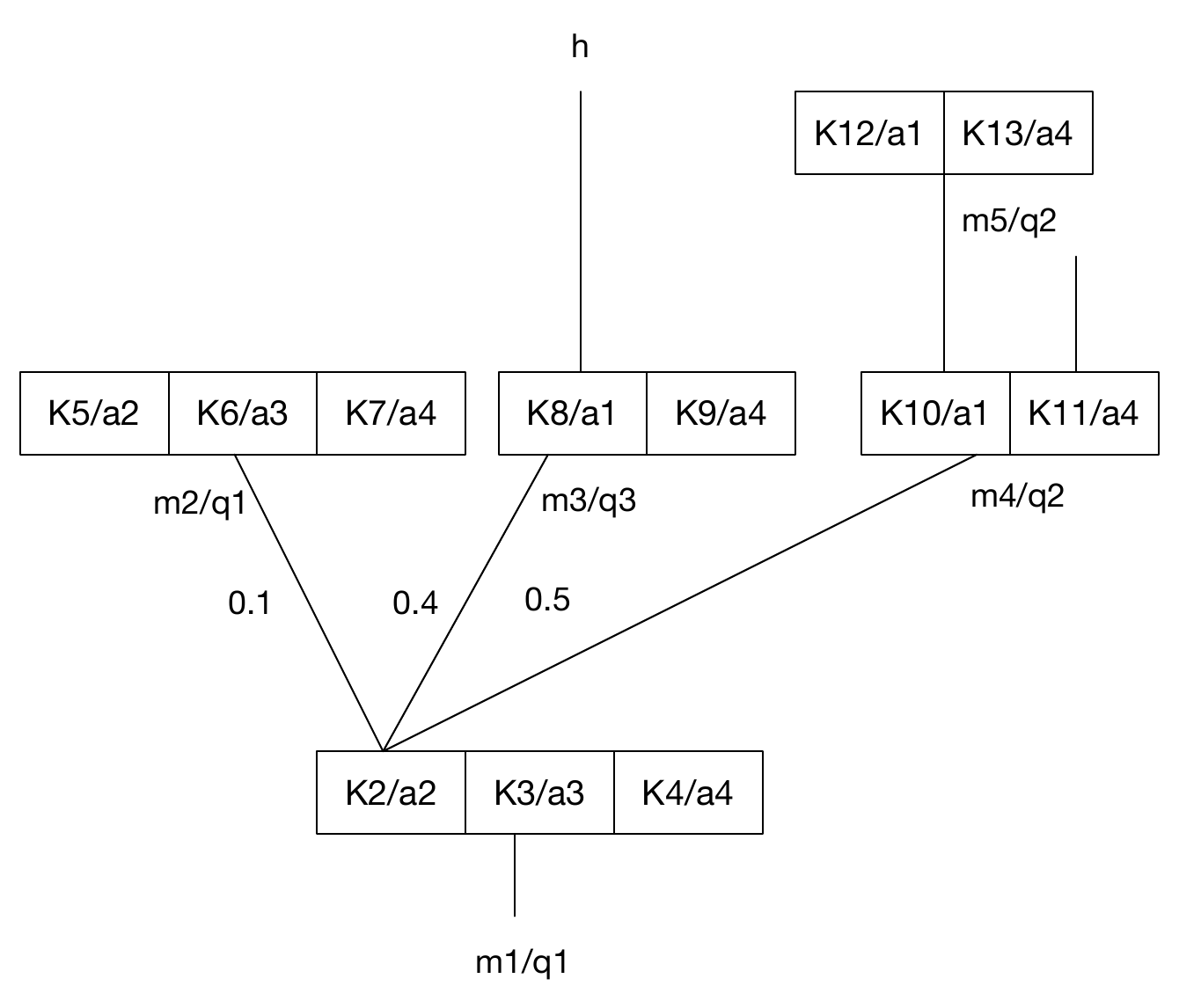}
        \caption{Partial stit model from MDP1.}
    \end{subfigure}
    \hfill
    \begin{subfigure}{0.45\textwidth}
        \centering
        \includegraphics[width=0.8\textwidth]{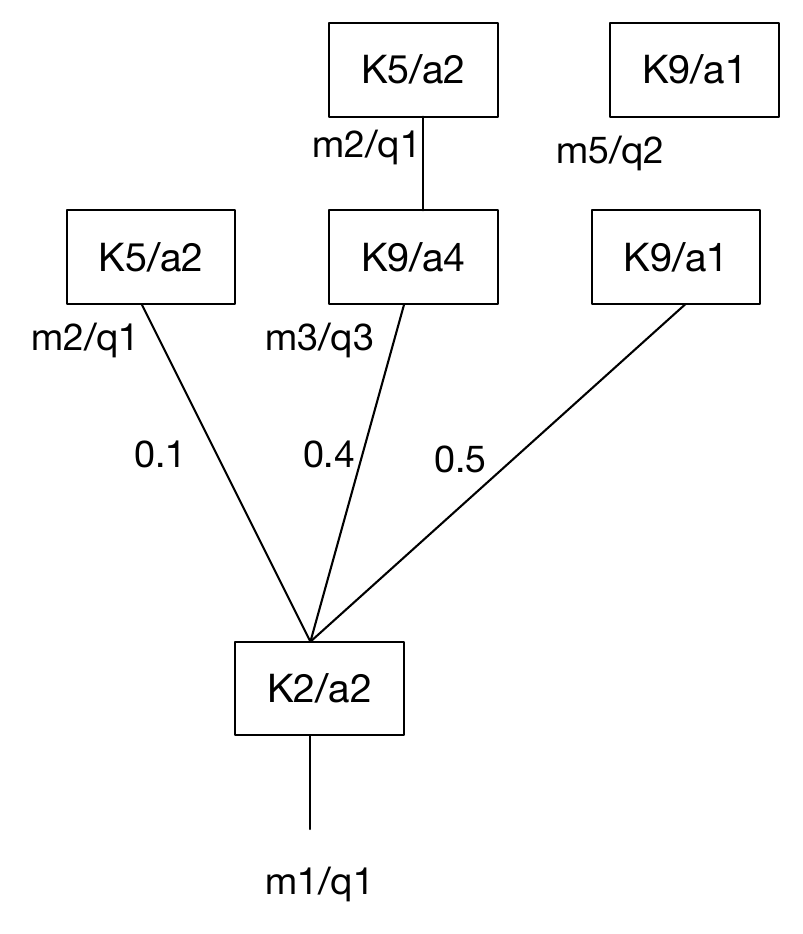}
        \caption{Partial stit model from MC in (b).}
    \end{subfigure}
    \caption{MDP, an induced MC, and corresponding stit model translations. Only part of the histories are shown. Each action is labeled with both the token name and MDP action, which can be modeled as a stit action type. Each moment is labeled with its name and the state the MDP is in at that moment. Numbers on histories indicate the probability of transitioning from $m$ to $m'$ under a given action token.}
    \label{fig:mdp_with_policy}
\end{figure}

\textbf{Semantics.}
Fix a Markov chain and state $s_0\in S$.  
Formulas $\top$, $p$, $\land$ and $\neg$ stand for the constant True, an atomic proposition, conjunction and negation respectively, with the usual semantics. 
$s_0\models P_J(\varphi)$ iff (if and only if) the probability mass of the set of paths starting at $s$ (so-called $s$-paths) and satisfying path formula $\varphi$ falls within the interval $J$. 
Given an $s_0$-path $\delta = s_0 s_1 s_2\dots \in S^\omega$ in the chain, $\delta \models X\pctlpath$ iff $s_1 s_2\dots \models \pctlpath$, $\delta \models \pctlstate_1 U \pctlstate_2$ (an \textit{Until} formula) iff $s_ks _{k+1}\dots \models \pctlstate_2$ for some $k\geq 0$ and $s_j s_{j+1}\dots \models \pctlstate_1$ for every $0\leq j<k$,
and $\delta \models  \pctlstate_1 U^{\leq n} \pctlstate_2$ if moreover $j \leq n$.
The Eventually or Future tense can be derived as $F\pctlstate \defeq \top U \pctlstate$ and the Always or Globally tense is its dual: $G\pctlstate := \neg F\neg \pctlstate$.

A \textit{Markov Decision Process} (MDP) is a tuple $\mdp=(S, \iota, AP, L, A, T, R)$ where $S,\iota, AP$ and $L$ are as before, $A$ is the finite set of actions that can be chosen by the agent acting in the MDP, $T:S\times A\times S\rightarrow [0,1]$ is the transition probability function s.t. for all states $s$ and actions $a$, $\sum_{s'\in S}T(s,a,s')\in \{0,1\}$, and $R:S\rightarrow \reals$ is the reward function.
We write $A(s)$ for the set of actions available to the agent at state $s$, and
$T(s, a, s')$ for the probability of reaching state $s'$ from $s$ by taking action $a \in A(s)$. 
Finally $R(s) \in \reals$ is the reward that the agent receives for entering state $s$.

A \textit{deterministic policy} $\pi:S\rightarrow A$ s.t. $\pi(s)\in A(s)$ represents an agent's decision making.\footnote{We don't consider stochastic policies, which map states to probability mass functions on $A$. In many cases where an optimal policy exists, a deterministic optimal policy exists.}
A policy resolves the non-determinism in the MDP by picking one and only one action at each state. Thus when following a policy, the MDP $\mdp$ becomes a Markov chain which we denote by $\mcmdpi$, and whose probability transition is given by $P(s,s')=\sum_{s'\in S}T(s,\pi(s),s')$.
Given a PCTL state formula $\pctlstate$, a policy $\pi$ is called \textit{feasible} if $s_{init}\models \pctlstate$ in the chain $\mcmdpi$. We write this as $s_{init} \models_\pi \pctlstate$.

In a Markov chain it is possible to compute the probability of satisfying a PCTL state formula. In an MDP that probability is undefined given the action non-determinism, but it is possible to compute the maximum and minimum probabilities of satisfying a path formula from any state. (Max and min are taken over all policies).
These are denoted by $P_{max}(\varphi)$ and $P_{min}(\varphi)$.

By construction, the agent's goal is to maximize its aggregate reward, or \textit{utility}. The most often used utility is the long-term expected discounted reward:  $V_\pi(s) \defeq \Exp_{\pi}[\sum_{t\geq 0}\gamma^tR(s_t)]$. 
Here, $\Exp_\pi$ is the expectation over all executions that start in $s$ and follow policy $\pi$ thereafter, and $\gamma\in (0,1)$ is a fixed discount factor. 
$V_\pi(s)$ is known as the \textit{value} of state $s$ under policy $\pi$.
Classical algorithms like value iteration \cite{sutton2018reinforcement} compute the \textit{Bellman-optimal value function} $V^*: S \rightarrow \reals$ and optimal policy $\pi^*$, defined by
\begin{equation}
    \label{eq:mdp optimal value fnt}
    \text{For all } s \in S,~V^*(s) = V_{\pi^*}(s) = \max_\pi V_\pi(s)
\end{equation}
Note that in this last equation the maximization is over all possible policies.

\section{Connection With STIT Logic}
Obligations (and permissions and prohibitions) of an RL agent are to be specified in stit logic, and the latter is interpreted over stit models, not MDPs. 
So verifying the normative compliance of an RL agent depends on a translation from MDPs to stit models (where the stit formulas can be evaluated).
In \cite{shea2022eau,SheaBlymyer_Abbas_2024} we provided just such a translation. 
By embedding deontic logic into the policy improvement process, we bridge utility-maximizing behavior with formal obligations. 

\subsection{MDP-to-Stit Translation}
The logic we use is Expected Act Utilitarianism, or EAU, which is based on dominance act utilitarianism \cite{horty2017action}.
For this paper's purposes we will not need the full set of definitions for EAU. The main theorem and its proof do not make use of the purely stit machinery. However, to understand the relation between the theorem and our overall purpose - normative compliance of RL agents - we will need to say a few words about the relation between RL and PCTL on the one hand, and stit models and EAU on the other. 
We cover the necessary EAU material here and defer the rest to the appendix for completeness.
\begin{definition}[Stit Model\cite{Horty01DLAgency}]
    \label{def:stit}
    Fix a finite set $\Agent$ of agent names. 
    A \emph{finite stit model} is a tuple $\Model = (\Tree,<,root,L, \Value, AP)$ where 
    \begin{itemize}
        \item $\Tree$ is a non-empty tree structure whose elements are called \textit{moments}, with strict partial order $<$, and with finite branching (so every moment has a finite number of successors). A \textit{history} $h$ is a maximal linearly ordered sequence of moments. $(m)^+_h$ is the immediate successor moment to $m$ on history $h$ (i.e. a child node of $m$ in the tree). 
        The set of all histories in $\Tree$ is $H$.
        Given moment $m$, the set of histories that pass through moment $m$ is $H_m$. $Ind \subset \Tree \times H$ is the set of pairs or \textit{indices} $m/h$ s.t. $m\in h$. 
        $root$ is the unique moment which is less than every other moment.
        \item $AP$ is a set of atomic propositions and $L:Ind\rightarrow 2^{AP}$ labels each index with the atomic propositions that hold there.
        \item $\Value: H \rightarrow \reals$ assigns a real-valued utility to each history.\footnote{The word `value' has two referents in this paper: $V_\pi(s)$ is the value of a state in an MDP and $\Value(h)$ is the value of a history in a stit model. We choose to overload the word in this way because it is so widely used in the RL and stit literatures that changing one of the uses is likely to increase confusion, not decrease it.}
        \item At moment $m$, $\Choiceam$ is a finite partition of $H_m$, whose cells represent the \textit{action tokens}, or actions, available to agent $\alpha$ at $m$. $\Choiceam$ satisfies
        \begin{itemize}
            \item No choice between undivided histories: for every two histories $h,h'$ in $H_m$, if there exists a future moment $m' > m$ that belongs to both $h$ and $h'$ then $h$ and $h'$ belong to exactly the same action tokens in $\Choiceam$.
            \item Independence of agency: $\Choiceam \cap \Choice{\beta}{m} \neq \emptyset$ for any two agents $\alpha,\beta$ .
        \end{itemize}
        Given $h \in H_m$, $\Choiceam(h)$ is the unique action token that contains it.

    \end{itemize}
    A \textit{strategy} is a partial map from moments to actions at those moments.
    Where there is no room for ambiguity, we drop $\alpha$ and/or $m$ from the notation.
\end{definition}

We illustrate the MDP-to-stit model translation process by example in Fig.~\ref{fig:mdp_with_policy}.
Given an MDP starting from some initial state $s_0$, one can unroll the MDP into a stit frame where $s_0$ is mapped to an initial moment $root$, successor MDP states are mapped to successor moments, and the actions available at every state are mapped to action tokens (with new names) available at the associated moments. 
Index $m/h$, translated from state $s$, is labeled with $L(s)$.
Because an MDP state, including $s_{init}$, can be visited multiple times in the course of an execution, it maps to multiple indices in the stit model, once per visit. 
An $s_{init}$-path is thus naturally translated to a history that starts at $root$, or at any of the translations of $s_{init}$.
It now makes sense to interpret a PCTL state formula at a given index $m/h$ \textit{in the stit model}, and a path formula along a history $h$.

The basic EAU obligation formula is written $\OEstit{\alpha}{\pctlstate}$, where $\pctlstate$ is a PCTL state formula. Evaluated at index $m/h$, it is read as `Agent $\alpha$ ought to see to it that $\pctlstate$ holds'.
In EAU, an agent's obligation at moment $m$ is to bring about those states of affairs that obtain under \textit{optimal action}, regardless of how the non-determinism (due to nature and other agents) resolves itself.
This is formalized by defining a (usually partial) order between action tokens at $m$ (i.e., the elements of $\Choiceam)$, and defining non-dominated actions as being optimal.
The ordering between actions, in turn, is based on the values of histories allowed by an action, i.e. the histories in action token $K$.
In the probabilistic setting that currently concerns us, an optimal action $K \in \Choiceam$ is one whose \textit{expected value} is greater than or equal to the expected value of any other action in $\Choiceam$. 
The expected value of $K$ is an average over the values of the histories in $K$.
To make these notions precise, the transition probabilities $T(s,a,s')$ and reward values $R(s)$ must be mapped from the MDP to the stit model.
This was done in In \cite{shea2022eau,SheaBlymyer_Abbas_2024}. For this paper we need only retain that the translation is such that
\begin{enumerate}
    \item an action is optimal in the stit model precisely if it is translated from a Bellman-optimal action in the MDP. Thus,
    \item a strategy is optimal in the stit model precisely if it is translated from a Bellman-optimal optimal policy $\pi^*$ in the MDP. Therefore, and since EAU obligations are based on optimal strategies,  
    \item an RL agent \textit{has} an obligation, formalized as an EAU formula $\OEstit{\alpha}{\pctlstate}$, if following the optimal policy yields a Markov chain that satisfies the obligation's PCTL content $\pctlstate$, and
    \item an RL agent \textit{meets} its obligation $\eauformula$ if it is indeed acting optimally.
\end{enumerate}

\subsection{Explicit and Implicit Rewards}
An agent's mission and its normative constraints are kept separate in our work. 
For example, our home care robot might have a reward $R$ that rewards higher those executions in which it accomplishes three tasks in a shorter amount of time: reminding the person in its care to take their medication, helping them get in and out of the shower, and preparing a simple breakfast.
Ordering these tasks so as to accomplish them quickest might clash, however, with the social expectation that on most days, breakfast follows the shower, and every day medication should follow breakfast except on fasting days, and it is permitted to let the person tarry a bit over her coffee.
While it is simple enough to design a reward to specify accomplishing some tasks quickly, designing a reward that strikes the above balance of obligations and permissions is full of pitfalls ~\cite{pan2022rewardmisspec,christiano17rlfh,ngo23alignmentDL}, and calls for a higher level of abstraction in specifying these obligations, permissions, and prohibitions. 
In this paper, EAU provides that abstraction.

Here it is helpful to explicitly distinguish between two rewards which appear in our setup, one explicit and the other implicit. 
The \textit{given, explicit} reward $R$ is that of the MDP, and it encodes the mission of the agent: it is designed such that a policy that maximizes the corresponding utility $\Exp_{\pi}[\sum_{t\geq 0}\gamma^tR(s_t)]$ accomplishes the mission. 
The \textit{implicit} reward, let's call it $R_d$ for deontic, is that associated to the EAU obligation $\eauformula$: as we saw in the previous section, in EAU, an agent ought to bring about those things that are true of all histories emanating from optimal action(s), where an action is optimal if it has maximum utility.
In EAU interpreted over MDPs it is natural to postulate that an action's utility is the (long-term expected discounted) reward of that action under some reward function $R_d:S \rightarrow \reals$.\footnote{Natural but not essential. For example van Berkel and Lyon introduce a stit deontic logic without utilities \cite{vanberkelL19neutralstit}.}
So by definition, satisfying $\eauformula$ means to act optimally according to $R_d$.
Thus finding a $\eauformula$-constrained policy that maximizes $V_\pi(s)$ is the same as finding a policy that maximizes $V_\pi(s)$ \textit{out of all policies that maximize $\Exp_\pi[\sum_t\gamma^tR_d(s_t)]$.}
This bi-level structure to the problem remains implicit in our approach since we do not actually have $R_d$.\footnote{Abarca and Broersen \cite{broersen21utilstit} introduce bi-valued non-stochastic stit models to analyze objective vs subjective obligations, but those were used for technical rather than conceptual reasons having to do with the proof of completeness of their logic. That said, it would be interesting to see if their models might shed any light on the bi-level structure we have here.}

Now there exist results asserting the impossibility of creating Markovian rewards to capture requirements in LTL \cite{yang2022reinforcement}, i.e. such that a reward-maximizing policy necessarily satisfies the LTL formula.
LTL and PCTL are not comparable in their expressiveness \cite{markey06essli}, so strictly speaking it is not known whether these results extend to PCTL formulas (though we suspect they do). 
If they do, then we are in the position that we have to analyze (the content of EAU) obligations as the propositions that hold in the best of all possible worlds, where `best' is according to a history-level utility that does not arise from aggregating single step rewards. 
\textit{A priori}, this does not pose philosophical difficulties (even if from a system design perspective it creates real practical difficulties; for instance it might mean having unbounded memory to track which execution the system is on to award it the right utility.)
But analytically, we now need a criterion for determining which histories get which utilities ... and this calls for a declarative or logical formalization of that criterion! 
This circularity, from `best' to `maximal utility' back to `best', seems to obviate the need for specifying a deontic utility altogether, something which goes against the basic premise of oughts in stit logics: namely, that obligations are derived from utilities.

We finally note that actions are explicit in MDPs, but are endogeneous or implicit in stit models.
This makes it that stit models with action types are a better translation of MDPs. 
A \emph{finite stit model with action types} \cite{horty2019epistemic} adds to the tuple of Def.~\ref{def:stit}, for each agent $\alpha$, a finite set $Type_\alpha$ of \textit{action types}. At every $m$, every action in $\Choiceam$ has an associated type from $Type_\alpha$.
Adding this to our translation is simple, and illustrated in Fig.~\ref{fig:mdp_with_policy}. Where state $s$ is visited by the agent at moment $m$, $Type_\alpha^m = A(s)$ and action $a\in A(s)$ translates to new action token $K$ of type $a$.
 
Through the correspondence between stit models and MDPs we are able to focus on the PCTL content of the obligation for model-checking and policy synthesis.
We can now move on to the algorithmic core of this paper.

\section{Policy Improvement in Deontically Constrained RL}
\label{sec:policy improvement}

\subsection{Policy Improvement Algorithm}
  
Recall that given a PCTL state formula $\pctlstate$, a feasible policy $\pi$ is one such that $s_{init}\models_\pi \pctlstate$.
Our problem is therefore
\begin{equation}
    \label{eq:core problem}
    \max_{\pi: s_{init}\models_\pi \pctlstate} V_\pi(s_0)
\end{equation}
Algorithm 1 is a combination of policy improvement \cite{sutton2018reinforcement} and PCTL model-checking procedures~\cite[Ch.~10]{baier2008principles}.\footnote{Detailed explanation provided in Appendix~\ref{appendix:new_pi constrained reach}.} 
It handles PCTL formulas of the form $P_{\geq \lambda}(\Phi  ~\mathbf{\until}~ \Psi)$, while complex PCTL formulas can be addressed by applying the standard recursive abstraction technique from PCTL model checking, where MDP states are recursively labeled by the sub-formulas they satisfy~\cite{baier2008principles}.
Given that $F\pctlstate \equiv \top ~\until ~\pctlstate$, a special case of Algorithm 1 is unconstrained reachability $P_{\geq \lambda}(\eventually b)$, where $b$ is the atomic proposition indicating membership in a set of target states $\mathbf{B}$ (i.e., $b \defeq s \in \mathbf{B}$). 
This is commonly abbreviated $P_{\geq \lambda}(\eventually~\mathbf{B})$.

Algorithm 1 proceeds as follows.
First a feasible policy $\pi^0$ is found by computing $P_{max}(\pctlpath)$, where $\pctlpath$ is a constrained reachability formula of the form $\Phi ~\until ~\Psi$. 

Then it is refined: at each state, we determine whether to change $\pi^t(s)$ to a new action (Step 6). 
The set of actions to choose from, $C^{t+1}(s)$, is not all of $A(s)$ as in regular policy improvement. Rather, only actions that maintain a sufficiently high probability of satisfaction $x_s^{(t+1)}$ are retained (Step 4). Here, $x_s^t$ denotes the probability of satisfying the PCTL path formula $\pctlpath$ starting from state $s$ while following policy $\pi^t$. To compute $x_s^t$, we first identify two key subsets of states: $S_{\text{yes}}$, the set of states from which $\pctlpath$ is satisfied with probability 1, and $S_{\text{no}}$, the set of states from which it is impossible to satisfy $\pctlpath$ (i.e., satisfaction probability is 0). For all $s \in S_{\text{yes}}$ we set $x_s = 1$, and for all $s \in S_{\text{no}}$, we set $x_s = 0$. We solve a linear system for the remaining states defined by the transition probabilities under $\pi^t$.
In the special case of unocnstrained reachability $P_{\geq \lambda}(\top~\until~ b)$, where $b \equiv s 
\in \mathbf{B}$, at least one action that satisfies PCTL constraint - that is, one included in $C^{t+1}(s)$ - exists if $\mathbf{B}$ is reachable from $s$, namely the current policy action $\pi^t(s)$ qualifies.
Otherwise, it doesn't matter what action is taken so the entire $A(s)$ is allowed.

The choice of action from $C^{t+1}(s)$ is $\epsilon$-greedy (Step 6.b): with probability $\epsilon$ a new action is chosen \textit{uniformly at random} from $C^{(t+1)}(s)$, and with probability $1-\epsilon$ the action that maximizes the $Q$-value is chosen. 
The $Q$ function gives the value of taking action $a$ at state $s$ then following policy $\pi$ afterwards: $Q(s,a) = R(s)+\Exp_{s'\sim T(s,a,\cdot)}[\gamma V_\pi(s')]$). 
In (Step 7), if $\epsilon=0$, the policy is updated only if the selected action improves utility.  
When $\epsilon > 0$, the selected action is applied unconditionally. 
Thus with positive $\epsilon$, the algorithm might accept a new action that actually decreases utility $\epsilon\%$ of the time (for each state), with the idea that perhaps this worse policy allows a later transition to a better policy. 
\begin{tcolorbox}[colback=gray!10, colframe=black, title=Algorithm 1: Policy Improvement with PCTL Constraints]
\label{algo:general_policy_improvement}
\scriptsize
\begin{enumerate}
    \item \textbf{Inputs:} MDP $\mdp$, PCTL formula $P_{\geq \lambda}(\Phi \, \until \, \Psi)$, $\epsilon \in [0,1)$

    \item \textbf{Initialization:} Set an initial feasible policy $\pi^0$ arbitrarily for all $s \in \mathcal{S}$. Initialize $V$.

    \item \textbf{Policy Refinement:} Repeat until the policy is stable:
    
    For each state $s \in \mathcal{S}$:
    \begin{itemize}
        \item \textbf{Step 1:} Evaluate the value function $V$ under the current policy $\pi^t$.

        \item \textbf{Step 2:} Compute the sets of states:
        \begin{itemize}
            \item For general formula $P_{\geq \lambda}(\Phi \, \until \, \Psi)$: 
            \[
            S_{no} = \texttt{Prob0}(Sat(\Phi), Sat(\Psi)), \quad S_{yes} = \texttt{Prob1}(Sat(\Phi), Sat(\Psi), S_{no}) \] 
            \item For the special case of a reachability formula $P_{\geq \lambda}(F~ b)$ with $b=s \in \mathbf{B}$ for some $B \subseteq S$: 
            \[
            S_{yes} = \mathbf{B}, \quad S_{no} = \{ s \mid \mathbf{B} \text{ unreachable from } s \}
            \] 
        \end{itemize}

        \item \textbf{Step 3:} Solve for vector $x_s$ of reachability probabilities:
        \[
        x_s^{(t+1)} = 
        \begin{cases} 
            1 & \text{if } s \in S_{yes}, \\
            0 & \text{if } s \in S_{no}, \\
            \sum_{s'} T(s, \pi^{t}(s), s') \cdot x_{s'}^{(t)} & \text{otherwise}
        \end{cases}
        \]

        \item \textbf{Step 4:} Compute reachability probabilities under change of action $a$:
        \[
        \text{For each action } a\in A(s),~Y^{(t+1)}(s,a) = \sum_{s'} T(s, a, s') \cdot x_{s'}
        \]

        \item \textbf{Step 5:} Determine valid actions:
        \[
        C^{(t+1)}(s) = 
        \begin{cases} 
            \{ a \in A(s) \mid Y^{(t+1)}(s,a) \geq \lambda \} & \text{if } s \notin S_{no}, \\
            A(s) & \text{otherwise}
        \end{cases}
        \]

        \item \textbf{Step 6:} For each $a \in C^{(t+1)}(s)$, compute its $Q$-value and:

        \begin{itemize}
            \item \textbf{(a)} Evaluate value function under temporary policy with $a$.
            \item \textbf{(b)} Select $a^*$ via $\epsilon$-greedy:
            \[
            a^* = 
            \begin{cases}
                \text{select uniformly at  random from } C^{(t+1)}(s) & \text{with probability } \epsilon, \\
                \arg\!\max\limits_{a \in C^{(t+1)}(s)} Q(s, a)& \text{with probability } 1 - \epsilon
            \end{cases}
            \]

        \end{itemize}

        \item \textbf{Step 7:} If $\epsilon = 0 $ (greedy case), update $\pi(s) \gets a^*$ only if it improves utility. 
Otherwise ($\epsilon > 0 $), update $\pi(s) \gets a^*$unconditionally.
    \end{itemize}

    \item \textbf{Final Value Function:} After convergence, compute $V_\pi$ via:
    \[
    V_{\pi}^{k+1}(s) = R(s) + \gamma \sum_{s'} T(s, \pi(s), s') V_{\pi}^k(s') \quad \text{until } \| V^{k+1} - V^k \| < \delta
    \]
\end{enumerate}
\end{tcolorbox}
 
This helps avoid getting stuck in local optima and improves performance in more complex environments.
The smaller $\epsilon$, the greedier the algorithm and the less exploration it performs.
We can now state a key property of our algorithm.
First, a definition.

\begin{definition}[Local Optimality]
A policy $\pi^*$ is a local optimizer of the constrained optimization problem \eqref{eq:core problem} if, for every state $s \in S$, switching $\pi^*(s)$ to any other valid action $a \in A(s)$ does not improve expected utility without violating the PCTL constraint.  
\end{definition}

\begin{theorem}
\label{thm:pi reach}
    With $\epsilon=0$, Algorithm 1 converges to a feasible policy that is a local maximizer of \eqref{eq:core problem}.
\end{theorem}
Intuitively, with zero $\epsilon$, the algorithm updates a policy only when this improves utility. As a result, it converges to policies that cannot be improved locally, even if a globally better policy exists. 
This outcome depends on the initial policy and the MDP's structure.

\begin{proof}
Let $\pi^t$ denote the policy at iteration $t$, and let $V_{\pi^t}(s)$ be the expected discounted return starting from state $s$ under policy $\pi^t$.

\textit{(1) Feasibility is preserved.}  
By construction, the initial policy $\pi^0$ is feasible: it satisfies the PCTL constraint $\psi$ (e.g., $P_{\geq \lambda}(\Phi ~\until~ \Psi)$).  
At each iteration, the set of valid actions $C^{(t+1)}(s)$ at each state $s$ contains only those actions that ensure continued satisfaction of the constraint (i.e., result in a high enough reachability probability). Therefore, any policy update made during the algorithm preserves feasibility. Thus, all $\pi^t$ satisfy $\psi$.

\textit{(2) Value improvement at each iteration.}  
For each state $s \in \mathcal{S}$, the algorithm updates $\pi^t(s)$ only if there exists a valid action $a \in C^{(t+1)}(s)$ such that the value $V(s)$ increases (i.e., $V_{\pi^{t+1}}(s) > V_{\pi^t}(s)$).  
This guarantees that the total expected utility is non-decreasing from iteration to iteration. Because $\epsilon = 0$, no exploratory (i.e., non-improving) updates are allowed.

\textit{(3) Finiteness of the policy space.}  
The number of deterministic policies is finite: at most $|\mathcal{A}|^{|\mathcal{S}|}$.  
Since each update improves the value of the policy at least at one state and no policy is repeated, the algorithm must terminate after a finite number of iterations.

\textit{(4) Convergence to a locally Bellman-optimal policy.}  
At termination, the policy $\pi^*$ is such that no action substitution at any state $s$ — within the set $C(s)$ of valid (constraint-preserving) actions — improves the value.  
Equivalently, $\pi^*$ satisfies the Bellman optimality condition restricted to the feasible set:
\[
V_{\pi^*}(s) = \max_{a \in C(s)} \left[ R(s) + \gamma \sum_{s'} T(s,a,s') V_{\pi^*}(s') \right]
\]
for all $s \in \mathcal{S}$.  
This ensures that $\pi^*$ is a locally Bellman-optimal policy in the space of feasible policies, i.e., a local optimum of the constrained objective.
\end{proof}
If $\epsilon>0$ then with probability 1, in the infinite iterations limit, the algorithm eventually visits a globally maximizing policy.

But this asymptotic result is not practically useful as it amounts to little more than a brute force approach.

\section{Experimental Results}
\label{sec:experiments}

We implemented our algorithm for EAU-constrained policy improvement.
This section illustrates their operation on sample MDPs. 

\subsection{Experimental Setup}
We use two MDPs: MDP1 is a smaller environment with fewer states and actions used to assess Algorithms 1 while MDP2 is a more complex environment that includes potential pitfalls where policies may get stuck in suboptimal loops. It is used to test our algorithm's ability to overcome local optima. 

We use two PCTL formulas: $\formula_R \defeq P_{\geq \lambda}(F ~\mathbf{B})$ is a reachability formula,  and $\formula_{CR} \defeq P_{\geq \lambda}(\Phi~\until~\Psi)$ is a general constrained reachability formula.

The evaluation criteria are
\begin{itemize}
\item Constraint Satisfaction: We verify whether the final policy indeed satisfies the PCTL formula.
\item Value Function Evaluation: The value of each policy is assessed using Bellman’s equation \cite{sutton2018reinforcement}.
\item Convergence Behavior: We evaluate how quickly the algorithm converges to an optimal policy.
\item Runtime Performance: The computational efficiency of different methods is compared.
\end{itemize}

\paragraph{Experimental Process}
The general process is as follows.
\begin{enumerate}
\item Obtain ground truth: for evaluation purposes, the global maximizer $\pi^*$ of Eq. \eqref{eq:core problem} is obtained by brute force (generate all policies, retain those that satisfy the formula, and out of those return the policy with highest value $V_\pi(s_{init})$).
\item Run Algorithm 1 with $\epsilon=0$. Call the returned policy $\pipo$.
\item Run Algorithm 1 with $\epsilon >0$ (for MDP2 only). Call the final policy $\pi^\epsilon$.
\end{enumerate}

\subsection{Experiments on Unconstrained Reachability}
In this section, we evaluate Algorithm 1 on $P_{\geq \lambda}(\eventually ~\mathbf{B})$. 
\paragraph{Results for MDP1}
MDP1 was shown in Fig. \ref{fig:mdp_with_policy}(a). The formula is \( P_{\geq 0.4} (\eventually ~ s_3) \).

A brute force search returns that $\pi^* = \{s_0 \mapsto a_1, s_1 \mapsto a_2, s_2 \mapsto a_4, s_3 \mapsto a_4\}$. 
The probability $x_s$ of reaching $s_3$ from state $s$ under this optimal policy is given by $x_s = 1.0$ for all $s\in S$, and the optimal value is $V^*(s_0) = 71.05$.

Algorithm 1 is run ($\epsilon = 0$). 
Table \ref{tab:policy_improvement_mdp1} shows its iterations, with action changes highlighted in yellow. In this case, it converges quickly in 4 steps.
But asserting convergence requires re-visiting all the states once, to determine that no action changes will happen, so the algorithm terminates in 9 steps. In a large MDP, visiting all states can be costly.

In this case, Algorithm 1 ($\epsilon = 0$) actually converges to the global optimizer $\pi^*$ but in general this is not guaranteed.

\begin{table}[t]
\caption{Policy updates during Algorithm 1 ($\epsilon = 0$) on MDP1 for formula $P_{\geq 0.4} ( \eventually  ~ s_3)$. Yellow highlights states where the policy changes in that iteration.}
\centering
\small
\begin{tabular}{|c|c|c|c|c|c|c|}
\hline
\textbf{Step} & \textbf{$s_0$} & \textbf{$s_1$} & \textbf{$s_2$} & \textbf{$s_3$} & \textbf{$V(s_0)$} & \textbf{ $x_s$} \\
\hline
$\pi^0$ & $a_1$ & $a_3$ & $a_1$ & $a_4$ & $69.98$ & $[0.44, 0.44, 0.00, 1.00]$ \\
$s_0 \rightarrow \pi^0$ & $a_1$ & $a_3$ & $a_1$ & $a_4$ & $69.98$ & $[0.44, 0.44, 0.00, 1.00]$ \\
$s_1 \rightarrow \pi^1$ & $a_1$ & \cellcolor{yellow!50}$a_2$ & $a_1$ & $a_4$ & $70.39$ & $[0.44, 0.44, 0.00, 1.00]$ \\
$s_2 \rightarrow \pi^2$ & $a_1$ & $a_2$ & \cellcolor{yellow!50}$a_4$ & $a_4$ & $71.05$ & $[1.00, 1.00, 1.00, 1.00]$ \\
$s_3 \rightarrow \pi^2$ & $a_1$ & $a_2$ & $a_4$ & $a_4$ & $71.05$ & $[1.00, 1.00, 1.00, 1.00]$ \\
$s_0 \rightarrow \pi^2$ & $a_1$ & $a_2$ & $a_4$ & $a_4$ & $71.05$ & $[1.00, 1.00, 1.00, 1.00]$ \\
$s_1 \rightarrow \pi^2$ & $a_1$ & $a_2$ & $a_4$ & $a_4$ & $71.05$ & $[1.00, 1.00, 1.00, 1.00]$ \\
$s_3 \rightarrow \pi^2$ & $a_1$ & $a_2$ & $a_4$ & $a_4$ & $71.05$ & $[1.00, 1.00, 1.00, 1.00]$ \\
$s_4 \rightarrow \pi^2$ & $a_1$ & $a_2$ & $a_4$ & $a_4$ & $71.05$ & $[1.00, 1.00, 1.00, 1.00]$ \\
\hline
\end{tabular}

\label{tab:policy_improvement_mdp1}
\end{table}

\subsubsection{Results for MDP2}
\begin{figure}[H]
    \centering
    \includegraphics[width=0.6\columnwidth]{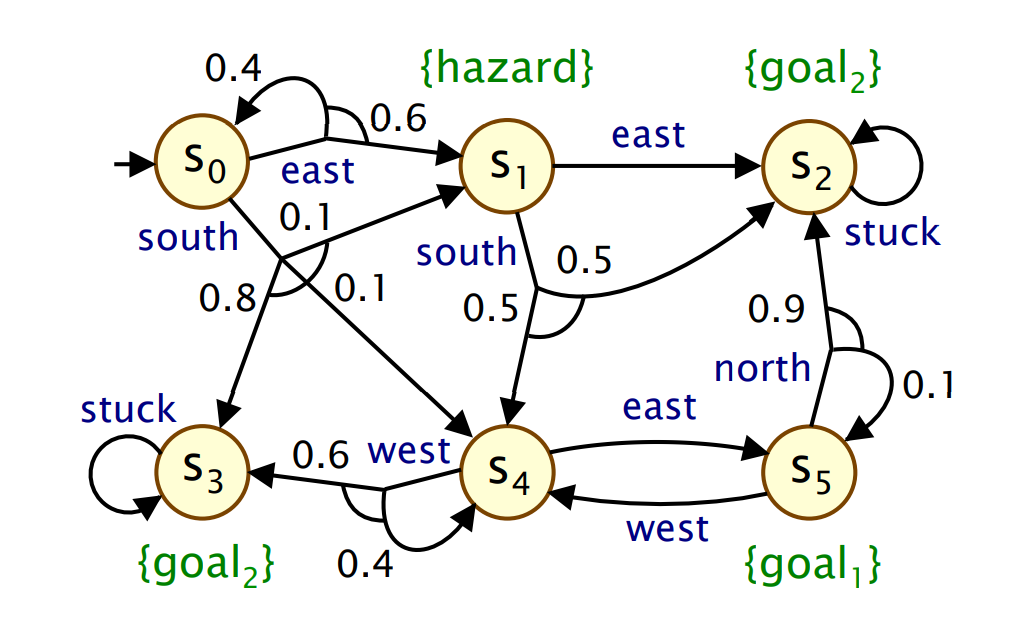}
    \caption{MDP2 from \cite{kwiatkowska2013automated}, which models a robot navigating a grid environment with probabilistic movement due to obstacles. The initial state is $s_0$. The actions are labeled east, west, south, north, and stuck. Atomic propositions are shown in green and between curly braces next to the labeled states, e.g., $L(s_1) = $ hazard. The rewards (not shown) are, in order of states, 1, 2, 3, 20, 0, 0.
}
    \label{fig:mdp2}
\end{figure}

MDP2 is shown in Fig. \ref{fig:mdp2}. The formula is $P_{\geq 0.3} ( \eventually  ~ s_2)$.
A brute force search returns two optimal policies depending on the action taken in $s_5$: $\pi^* = \{s_0 \mapsto east, s_1 \mapsto south, s_2 \mapsto stuck, s_3 \mapsto stuck, s_4 \mapsto west, s_5 \mapsto west \text{ or } north\}$. 
The vector $x$, whose element $x_s$ is the probability of reaching target $s_2$ from state $s$ under either optimal policy is given by $x = [0.50, 0.50, 1.00, 0.00, 0.00, 0.00]$. The optimal value function at the initial state is $V^*(s_0) = 78.71$.\\
Algorithm 1 is run ($\epsilon = 0$). 
In this case, a local maximum is quickly reached in 12 steps.
(Table~\ref{tab:policy_improvement_mdp2_local} in the Appendix shows its iterations). 
In this case, the algorithm converges prematurely to a local optimizer. While the final policy still satisfies the reachability constraint, it does not globally maximize the utility function.

\begin{figure}[H]
    \centering
    \includegraphics[width=0.65\columnwidth]{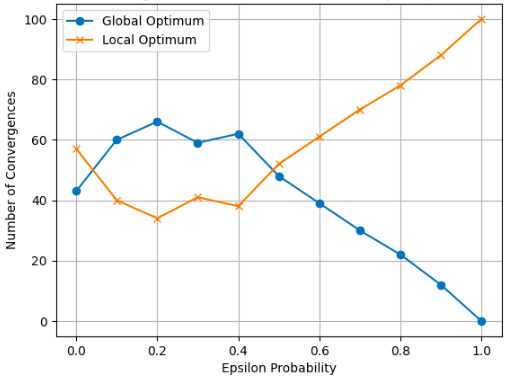}
    \caption{Empirical convergence of Algorithm 1 over 100 trials for each $\epsilon$, using MDP1 for the unconstrained reachability formula. Convergence is declared when a full iteration over all states results in no change to the policy—i.e., after one stable pass. Even though a larger $\epsilon$ allows more exploration, exploration's benefits only appear after a sufficiently large number of iterations, which is not the case here. Thus in practice, neither a very large nor a very small $\epsilon$ are used (which still leaves a lot of room for heuristic tuning, including a variable $\epsilon$).}
    \label{fig:greedy}
\end{figure}

\paragraph{Analysis of Exploration Effectiveness}
Figure \ref{fig:greedy} presents the empirical convergence results of Algorithm 1. We sweep $\epsilon$ between 0 and 1, and run Algorithm 1 a hundred times for each value, tracking how many runs reach a local vs. a global optimizer.  Here, we declare convergence when the policy remains unchanged over a complete outer iteration. 
For smaller values of $\epsilon$, the algorithm does not explore enough and often gets trapped in local optima.
For larger values of $\epsilon$, excessive random exploration reduces policy stability, preventing convergence altogether.
The balance between exploration and exploitation in this case is best achieved at $\epsilon$ = 0.1  \text{--} 0.4, where exploration is sufficient to escape local minima without causing too much instability.

\subsection{Experiments on Constrained Reachability}
\paragraph{Results for MDP1}
The formula is \(\eauformula = P_{\geq 0.56} (  \neg s_3  ~\until  ~s_2) \).
A brute force search returns that $\pi^* = \{s_0 \mapsto a_1, s_1 \mapsto a_3, s_2 \mapsto a_1, s_3 \mapsto a_4\}$. 
The corresponding probability vector $x$ of satisfying $\eauformula$ under this optimal policy is given by $x = [0.56, 0.56, 1.00, 0.00]$. The optimal value of the initial state is $V^*(s_0) = 69.98$.

Algorithm 1 ($\epsilon = 0$) is run. 
In this case, the global maximum is quickly reached in 8 steps.

\subsubsection{Results for MDP2}
The formula is $\eauformula_2 = P_{\geq 0.85} ( \neg hazard ~ \until ~ goal_2)$

A brute force search returns that $\pi^* = \{s_0 \mapsto south, s_1 \mapsto south, s_2 \mapsto stuck, s_3 \mapsto stuck, s_4 \mapsto west, s_5 \mapsto \{west, north\}\}$. 
The probability vector $x$ of satisfying $\eauformula_2$ under this optimal policy is given by $x= [0.9, 0.0, 1.0, 1.0, 1.0, 1.0]  $. The optimal value function at the initial state is $V^*(s_0) = 168.42$.
Algorithm 1 is run ($\epsilon = 0$). In this case, a local maximum is quickly reached in 13 steps.

Applying $\epsilon$-greedy exploration again allows the algorithm to escape local optima and reach the global optimum, as shown in Table \ref{tab:policy_improvement_mdp2_global} (in the Appendix). This confirms that controlled exploration is essential for improving performance across different PCTL constraints.

\begin{figure}[H]
    \centering
    \includegraphics[width=0.6\columnwidth]{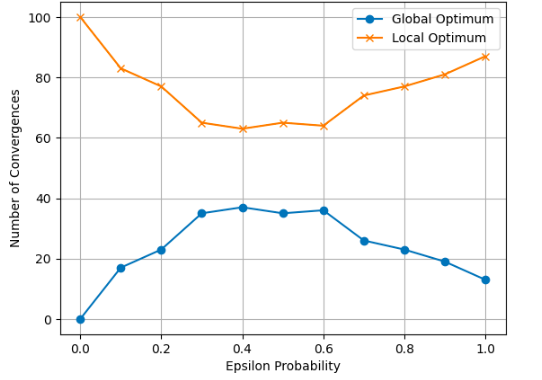}
    \caption{Convergence results over 100 trials with varying $\epsilon$ using MDP2 for constrained reachability formula. Keeping $\epsilon$ at 0.4 reduces the number of policies that get stuck in local optima.}
    \label{fig:greedy1}
\end{figure}

Figure \ref{fig:greedy1} presents the results of 100 independent runs of Algorithm 1 on MDP2, each with a different $\epsilon$ value. We recorded whether the final policy reached a global or local optimum.
Unlike the reachability case, the constrained reachability formula consistently resulted in fewer global optima across all $\epsilon$ values. The highest success rate in reaching the global optimum, 42 out of 100 trials, occurred at $\epsilon = 0.4$.

\subsection{Performance Analysis}

All experiments were conducted on a MacBook Pro with an Apple M1 processor (8-core CPU, 4 performance and 4 efficiency cores), 16GB of unified memory, and macOS Ventura 13.4. Code was implemented in Python 3.10.
The runtime comparison shows that the reachability case for MDP1 takes 0.1365 seconds, while for MDP2, it is 0.0482 seconds. MDP1 requires 0.1549 seconds for the constrained reachability case, whereas MDP2 completes in 0.0666 seconds, indicating that MDP2 consistently runs faster than MDP1.

\label{sec:conclusion}
\section{Conclusion}
This paper introduces a policy improvement algorithm for reinforcement learning agents that must satisfy deontic constraints expressed in PCTL. 

Our approach ensures that the agent maximizes utility while adhering to acceptable behavior constraints. We proved that our algorithm converges to a locally optimal policy within a finite amount of steps. Although our algorithm provably converges to such policies, they may not globally maximize expected utility. This reflects the standard limitations of local policy improvement in reinforcement learning.
Extending our framework to guarantee global optimality remains an open challenge.
Another avenue for future work is to extend our methods to partially observable MDPs.

\bibliographystyle{deon16}
\bibliography{deon25}

\begin{thebibliography}{10}
\expandafter\ifx\csname url\endcsname\relax
  \def\url#1{\texttt{#1}}\fi
\expandafter\ifx\csname urlprefix\endcsname\relax\def\urlprefix{URL }\fi
\newcommand{\enquote}[1]{``#1''}

\bibitem{broersen21utilstit}
Abarca, A. I.~R. and J.~M. Broersen, \emph{A deontic stit logic based on beliefs and expected utility}, in: J.~Y. Halpern and A.~Perea, editors, \emph{Proceedings Eighteenth Conference on Theoretical Aspects of Rationality and Knowledge, {TARK} 2021, Beijing, China, June 25-27, 2021},  {EPTCS}  \textbf{335}, 2021, pp. 281--294.
\newline\urlprefix\url{https://doi.org/10.4204/EPTCS.335.27}

\bibitem{Abel2016ReinforcementLA}
Abel, D., J.~MacGlashan and M.~L. Littman, \emph{Reinforcement learning as a framework for ethical decision making}, in: \emph{AAAI Workshop: AI, Ethics, and Society}, 2016.
\newline\urlprefix\url{https://api.semanticscholar.org/CorpusID:14717578}

\bibitem{achiam2017constrained}
Achiam, J., D.~Held, A.~Tamar and P.~Abbeel, \emph{Constrained policy optimization}, ICML  (2017).

\bibitem{eitan04cmdps}
ALTMAN, E., editor, \enquote{CONSTRAINED MARKOV DECISION PROCESSES,} Taylor \& Francis, 2004.

\bibitem{Alur2022transformingspecs}
Alur, R., S.~Bansal, O.~Bastani and K.~Jothimurugan, \enquote{A Framework for Transforming Specifications in Reinforcement Learning,} Springer Nature Switzerland, Cham, 2022 pp. 604--624.
\newline\urlprefix\url{https://doi.org/10.1007/978-3-031-22337-2_29}

\bibitem{baier2008principles}
Baier, C. and J.-P. Katoen, \enquote{Principles of model checking,} MIT press, 2008.

\bibitem{BROERSEN13probabilisticstit}
Broersen, J., \emph{Probabilistic stit logic and its decomposition}, International Journal of Approximate Reasoning \textbf{54} (2013), pp.~467--477, eleventh European Conference on Symbolic and Quantitative Approaches to Reasoning with Uncertainty (ECSQARU 2011).
\newline\urlprefix\url{https://www.sciencedirect.com/science/article/pii/S0888613X1200148X}

\bibitem{BROERSEN06stitCL}
Broersen, J., A.~Herzig and N.~Troquard, \emph{From coalition logic to stit}, Electronic Notes in Theoretical Computer Science \textbf{157} (2006), pp.~23--35, proceedings of the Third International Workshop on Logic and Communication in Multi-Agent Systems (LCMAS 2005).
\newline\urlprefix\url{https://www.sciencedirect.com/science/article/pii/S1571066106003197}

\bibitem{broersen06stitatl}
Broersen, J., A.~Herzig and N.~Troquard, \emph{A stit-extension of atl}, in: M.~Fisher, W.~van~der Hoek, B.~Konev and A.~Lisitsa, editors, \emph{Logics in Artificial Intelligence} (2006), pp. 69--81.

\bibitem{christiano17rlfh}
Christiano, P.~F., J.~Leike, T.~B. Brown, M.~Martic, S.~Legg and D.~Amodei, \emph{Deep reinforcement learning from human preferences}, in: \emph{Proceedings of the 31st International Conference on Neural Information Processing Systems}, NIPS'17 (2017), p. 4302–4310.

\bibitem{djeumou23irl}
Djeumou, F., C.~Ellis, M.~Cubuktepe, C.~Lennon and U.~Topcu, \emph{Task-guided irl in pomdps that scales}, Artificial Intelligence \textbf{317} (2023), p.~103856.
\newline\urlprefix\url{https://www.sciencedirect.com/science/article/pii/S0004370223000024}

\bibitem{Horty01DLAgency}
Horty, J., \enquote{Agency and Deontic Logic,} Cambridge University Press, 2001.

\bibitem{horty2019epistemic}
Horty, J., \emph{Epistemic oughts in stit semantics}, Ergo \textbf{6} (2019).

\bibitem{horty2017action}
Horty, J. and E.~Pacuit, \emph{Action types in stit semantics}, The Review of Symbolic Logic \textbf{10} (2017), pp.~617--637.

\bibitem{pautomata10}
Huth, M., N.~Piterman and D.~Wagner, \emph{p-automata: New foundations for discrete-time probabilistic verification}, in: \emph{2010 Seventh International Conference on the Quantitative Evaluation of Systems}, 2010, pp. 161--170.

\bibitem{Kasenberg_Scheutz_2018}
Kasenberg, D. and M.~Scheutz, \emph{Norm conflict resolution in stochastic domains}, Proceedings of the AAAI Conference on Artificial Intelligence \textbf{32} (2018).
\newline\urlprefix\url{https://ojs.aaai.org/index.php/AAAI/article/view/11295}

\bibitem{kwiatkowska2013automated}
Kwiatkowska, M. and D.~Parker, \emph{Automated verification and strategy synthesis for probabilistic systems}, in: \emph{Proceedings of the International Symposium on Automated Technology for Verification and Analysis (ATVA)} (2013), pp. 5--22.

\bibitem{morteza2012}
Lahijanian, M., S.~B. Andersson and C.~Belta, \emph{Temporal logic motion planning and control with probabilistic satisfaction guarantees}, IEEE Transactions on Robotics  (2012).

\bibitem{rlj2024omega}
Lindemann, L., N.~Jansen and U.~Topcu, \emph{Optimizing rewards while meeting omega-regular constraints}, in: \emph{Proceedings of the 38th AAAI Conference on Artificial Intelligence}, 2024.

\bibitem{markey06essli}
Markey, N., \emph{Expressiveness of temporal logics} (2006).
\newline\urlprefix\url{http://www.lsv.fr/~markey/Teaching/ESSLLI06/ESSLLI-proc.pdf}

\bibitem{topcu21rewardmachines}
Neary, C., Z.~Xu, B.~Wu and U.~Topcu, \emph{Reward machines for cooperative multi-agent reinforcement learning}, in: \emph{Proceedings of the 20th International Conference on Autonomous Agents and MultiAgent Systems}, AAMAS '21 (2021), p. 934–942.

\bibitem{ngo23alignmentDL}
Ngo, R., L.~Chan and S.~Mindermann, \emph{The alignment problem from a deep learning perspective} (2022), v4.
\newline\urlprefix\url{https://arxiv.org/pdf/2209.00626.pdf}

\bibitem{pan2022rewardmisspec}
Pan, A., K.~Bhatia and J.~Steinhardt, \emph{The effects of reward misspecification: Mapping and mitigating misaligned models}, in: \emph{International Conference on Learning Representations}, 2022.
\newline\urlprefix\url{https://openreview.net/forum?id=JYtwGwIL7ye}

\bibitem{ropke2024divide}
R{\"o}pke, W., M.~Reymond, P.~Mannion, D.~M. Roijers, A.~Nowe and R.~R{\u{a}}dulescu, \emph{Divide and conquer: Provably unveiling the pareto front with multi-objective reinforcement learning}, in: \emph{Seventeenth European Workshop on Reinforcement Learning}, 2024.
\newline\urlprefix\url{https://openreview.net/forum?id=EruoSfUPxU}

\bibitem{shea2022eau}
Shea-Blymyer, C. and H.~Abbas, \emph{Generating deontic obligations from utility-maximizing systems}, in: \emph{Proceedings of the 2022 AAAI/ACM Conference on AI, Ethics, and Society}, 2022, pp. 653--663.

\bibitem{SheaBlymyer_Abbas_2024}
Shea-Blymyer, C. and H.~Abbas, \emph{Formal ethical obligations in reinforcement learning agents: Verification and policy updates}, Proceedings of the AAAI/ACM Conference on AI, Ethics, and Society \textbf{7} (2024), pp.~1368--1378.
\newline\urlprefix\url{https://ojs.aaai.org/index.php/AIES/article/view/31730}

\bibitem{skalse2022rewardhypothesis}
Skalse, J. and A.~Abate, \emph{The reward hypothesis is false}, NeurIPS  (2022).

\bibitem{sutton2018reinforcement}
Sutton, R.~S. and A.~G. Barto, \enquote{Reinforcement Learning: An Introduction,} MIT Press, 2018, 2nd edition.

\bibitem{vanberkelL19neutralstit}
van Berkel, K. and T.~Lyon, \emph{A neutral temporal deontic stit logic}, in: \emph{Logic, Rationality, and Interaction: 7th International Workshop, LORI 2019, Chongqing, China, October 18–21, 2019, Proceedings} (2019), p. 340–354.
\newline\urlprefix\url{https://doi.org/10.1007/978-3-662-60292-8_25}

\bibitem{yang2022reinforcement}
Yang, C., M.~Littman and M.~Carbin, \emph{Reinforcement learning with general {LTL} objectives is intractable}, in: \emph{Combining Learning and Reasoning: Programming Languages, Formalisms, and Representations}, 2022.

\end{thebibliography}

\Appendix
\section{Computing Prob0 and Prob1 in Algorithm 1}
\label{appendix:new_pi constrained reach}

Algorithm \texttt{Prob0}(Sat($\Phi$), Sat($\Psi$)) identifies the set of states for which the probability of satisfying the constrained reachability formula $\Phi~ \until~ \Psi$ is zero. The algorithm begins with states that do not satisfy $\Phi$ or $\Psi$. It iteratively adds any state that satisfies $\Phi$ but can only transition (with positive probability) to states already known to be unsatisfiable. 
Once this set stabilizes, it represents the set of states from which reaching a $\Psi$-satisfiable state while remaining in $\Phi$-satisfiable state is impossible, and thus the probability of satisfying $\Phi ~\until ~\Psi$ is 0.

Algorithm $\texttt{Prob1}(Sat(\Phi), Sat(\Psi), S_{no})$ computes the set of states from which the formula $\Phi ~\until ~\Psi$ is satisfied with probability 1. It starts from the set $S_{no}$ of states known to violate the formula - that is, the states from which it is impossible to reach any state satisfying $\Psi$ while only visiting states that satisfy $\Phi$. These include states that are entirely disconnected from all $\Psi$-satisfying states and those from which any such path would have to pass through states that do not satisfy $\Phi$. The algorithm iteratively expands this set by adding any state that satisfies $\Phi$ that can only transition to other failure states (i.e., states in $S_{no}$). The complement of this expanded failure set yields the states from which the formula $\Phi ~\until~\Psi$ is guaranteed to hold. 

In the special case of an unconstrained reachability formula, $P_{\geq \lambda}(\eventually~\mathbf{B})$, this simplifies to the set $S_{no}$ which consists of all states from which $\mathbf{B}$ is unreachable.

\section{MDPs and Stit Frames}
\label{sec:appendix mdps stit}
We give some more details on our prior construction of the value function in stit frames for expected act utilitarianism.

\textbf{Probability measure over histories.}
Given a finite prefix $\hat{h}$ of a history starting at $m$,\footnote{A history starts at $m$ if its smallest moment is $m$.} its cylinder set is $C(\hat{h}) = \{h \in H_{m}~|~\hat{h} \text{ is a prefix of } h\}$. The $\sigma$-algebra associated with the stit frame is the smallest $\sigma$-algebra that contains all cylinder sets $C(\hat{h})$ where $\hat{h}$ ranges over all finite prefixes.
Given a set $L \subset H_m$ of histories, $\Proba(L|m)$ denotes the probability measure of $L$.

\textbf{Optimal actions.}
Let $\Utilityam(h)$ be equal to $\Value(h) \times \Proba(h | m)$. 

Then, in a moment $m$ such that \textit{either}
\begin{itemize}
    \item $m$ has no succeeding moment $m'$ where $|\Choice{\alpha}{m'}|>1$, \textit{or}
    \item $m$ has no succeeding moment $m'$ whose $\Choice{\alpha}{m'}$ contains two different actions $K,K'$ such that $\sum_{h \in K} \Utility{\alpha}{m'}(h) \neq \sum_{h' \in K'} \Utility{\alpha}{m'}(h')$
\end{itemize}
we take the \textit{quality} of an action $Q(K)$ as $\sum_{h \in K} \Utilityam(h)$ --- the sum of the utilities of its composing histories.
The first case handles models with end states (like finite games), or absorbing states.
The second case handles models with states where future choices don't change the available utility.
The latter is useful when $\Value$ is discounted sum, and in practice an $\epsilon$ difference may be assumed between available utilities.

In a moment $m$ that \textit{doesn't} meet the above criteria, we define $Q(K)$ recursively.
Let $(m)^+_{K} = \{(m)^+_h | h \in K\}$. Then 
\begin{equation}
    \label{eq:quality}
    Q(K) = \sum_{m' \in (m)^+_K} \Proba(m' | m) \max_{K' \in \Choice{\alpha}{m'}} Q(K')
\end{equation}

This means an action $K$'s quality is determined by the quality of the best action $K'$ in each of the moments $m'$ that $K$ leads to, modified by the probability of ending up in each moment $m'$ after taking the action.
An optimal action at moment $m$ is thus an action whose quality is not less than the quality of any other action at $m$.

\section{Policy Improvement Iterations}
The following tables illustrate how Algorithm 1 behaves on MDP2 with and without exploration. The first case demonstrates the unconstrained reachability formula $P_{\geq 0.3} ( \eventually ~ s_2)$ without exploration, where the algorithm converges to a local optimum. In contrast, the second case introduces $\epsilon$ - greedy exploration for the constrained reachability formula  $P_{\geq 0.85} ( \neg hazard ~ \until ~ goal_2)$, allowing the algorithm to escape the local optimum and to discover the globally optimal policy.  \\
To improve readability, we present only the iterations that include a policy change; steps with no updates are omitted.
\subsection{MDP2 Algorithm 1}

\label{appendix_label_a1_mdp2}
\begin{table}[H]
\caption{Results of Algorithm 1 ($\epsilon = 0$) on MDP2 without exploration for the reachability formula \( P_{\geq 0.3} ( \eventually ~ s_2)\) . Yellow highlights indicate the states where the policy changes. The algorithm converges to a local optimum. Steps after the final change are omitted.}
\centering
\small
\begin{tabular}{|c|c|c|c|c|c|c|c|}
\hline
\textbf{Step} & \textbf{$s_0$} & \textbf{$s_1$} & \textbf{$s_2$} & \textbf{$s_3$} & \textbf{$s_4$} & \textbf{$s_5$} & \textbf{$V(s_0)$} \\
\hline
$\pi^0$ & east & south & stuck & stuck & east & west & $14.64$  \\
$s_0 \rightarrow \pi^0$ & east & south & stuck & stuck & east & west & $14.64$ \\
$s_1 \rightarrow \pi^1$ & east & \cellcolor{yellow!50}east & stuck & stuck & east & west & $26.03$  \\
$s_2 \rightarrow \pi^1$ & east & east & stuck & stuck & east & west & $26.03$  \\
$s_3 \rightarrow \pi^1$ & east & east & stuck & stuck & east & west & $26.03$ \\
$s_4 \rightarrow \pi^1$ & east & east & stuck & stuck & east & west & $26.03$  \\
$s_5 \rightarrow \pi^2$ & east & east & stuck & stuck & east & \cellcolor{yellow!50}north & $26.03$  \\

\hline
\end{tabular}
\label{tab:policy_improvement_mdp2_local}
\end{table}

\subsection{MDP2 Algorithm 1 global optimizer}
\begin{table}[H]

\caption{Results of Algorithm 1 ($\epsilon > 0$) on MDP2  with $\epsilon$-greedy exploration for the constrained reachability formula \( P_{\geq 0.85} ( \neg hazard ~ \until ~ goal_2)\). Yellow highlights indicate the states where the policy changes. The algorithm escapes the local optimum and finds the globally optimal policy. Steps after the final change are omitted.}
\centering
\small
\begin{tabular}{|c|c|c|c|c|c|c|c|}
\hline
\textbf{Step} & \textbf{$s_0$} & \textbf{$s_1$} & \textbf{$s_2$} & \textbf{$s_3$} & \textbf{$s_4$} & \textbf{$s_5$} & \textbf{$V(s_0)$}  \\
\hline
$\pi^0$ & south & east & stuck & stuck & east & north & $149.71$  \\
$s_0 \rightarrow \pi^0$ & south & east & stuck & stuck & east & north & $149.71$  \\
$s_1 \rightarrow \pi^1$ & south & \cellcolor{yellow!50}south & stuck & stuck & east &  north & $149.53$  \\
$s_2 \rightarrow \pi^1$ & south & south & stuck & stuck & east &  north & $149.53$ \\
$s_3 \rightarrow \pi^1$ & south & south & stuck & stuck & east &  north & $149.53$  \\
$s_4 \rightarrow \pi^2$ & south & south & stuck & stuck & \cellcolor{yellow!50}west &  north & $168.42$  \\
$s_5 \rightarrow \pi^3$ & south & south & stuck & stuck & west &  \cellcolor{yellow!50}west & $168.42$  \\
\hline
\end{tabular}

\label{tab:policy_improvement_mdp2_global}
\end{table}

\end{document}